\documentclass[letterpaper, 10 pt, conference]{ieeeconf}  

\IEEEoverridecommandlockouts                              

\usepackage{graphics} 
\usepackage{epsfig} 
\usepackage{mathptmx} 
\usepackage{times} 
\usepackage{amsmath} 
\usepackage{amssymb}  

\usepackage{booktabs} 
\usepackage{multirow}
\usepackage{tikz} 
\usepackage{contour}
\usepackage{caption}
\usepackage{subcaption}
\usepackage{placeins}

\newcommand{\pent}{$p_{\mathcal{H}}$ }  

\newcommand{\any}{\contour{black}{\textcolor{white}{any}}}
\definecolor{car}{RGB}{100, 150, 245}
\definecolor{bicycle}{RGB}{100, 230, 245}
\definecolor{motorcycle}{RGB}{30, 60, 150}
\definecolor{truck}{RGB}{180, 30, 80}
\definecolor{othervehicle}{RGB}{0, 0, 255}
\definecolor{person}{RGB}{255, 30, 30}
\definecolor{motorcyclist}{RGB}{150, 30, 90}
\definecolor{bicyclist}{RGB}{100, 0, 50}
\definecolor{road}{RGB}{255, 0, 255}
\definecolor{parking}{RGB}{255, 150, 255}
\definecolor{sidewalk}{RGB}{75, 0, 75}
\definecolor{otherground}{RGB}{175, 0, 75}
\definecolor{building}{RGB}{255, 200, 0}
\definecolor{fence}{RGB}{255, 120, 50}
\definecolor{vegetation}{RGB}{0, 175, 0}
\definecolor{trunk}{RGB}{135, 60, 0}
\definecolor{terrain}{RGB}{150, 240, 80}
\definecolor{pole}{RGB}{255, 240, 150}
\definecolor{trafficsign}{RGB}{255, 0, 0}
\definecolor{ground}{RGB}{193, 38, 143}
\definecolor{structure}{RGB}{255, 160, 25}
\definecolor{nature}{RGB}{109, 179, 40}
\definecolor{objects}{RGB}{255, 120, 75}
\definecolor{vehicle}{RGB}{38, 55, 216}
\definecolor{human}{RGB}{164, 23, 65}
\definecolor{static}{RGB}{211, 130, 69}
\definecolor{dynamic}{RGB}{101, 39, 141}
\definecolor{any}{RGB}{255,255,255}
\newlength{\DepthReference}
\newlength{\HeightReference}
\newlength{\Width}%
\newcommand{\textb}[1]%
{%
	\settowidth{\Width}{#1}%
	\colorbox{#1}%
	{%
		\raisebox{-\DepthReference}%
		{%
			\parbox[b][\HeightReference+\DepthReference][c]{\Width}{\centering#1}%
		}%
	}%
}
\newcommand{\textd}[1]{\textcolor{#1}{#1}}

\newcommand{\class}[1]{\textit{#1}}

\usepackage[pagebackref,breaklinks,colorlinks]{hyperref}

\title{\LARGE \bf
Hierarchical Insights:\\Exploiting Structural Similarities for Reliable 3D Semantic Segmentation
}

\author{Mariella Dreissig$^{1, 2}$, Florian Piewak$^{1}$ and Joschka Boedecker$^{2}$
\thanks{This work has been submitted to the IEEE for possible publication. Copyright may be transferred without notice, after which this version may no longer be accessible.}%
\thanks{$^{1}$ RD Mercedes-Benz AG, Sindelfingen, Germany. Main contact:
        {\tt\small mariella.dreissig@mercedes-benz.com}}%
\thanks{$^{2}$University of Freiburg, Freiburg, Germany.}%
}

\begin{document}

\maketitle
\thispagestyle{empty}
\pagestyle{empty}

\begin{abstract}
	Safety-critical applications such as autonomous driving require robust 3D environment perception algorithms capable of handling diverse and ambiguous surroundings. The predictive performance of classification models is heavily influenced by the dataset and the prior knowledge provided by the annotated labels. While labels guide the learning process, they often fail to capture the inherent relationships between classes that are naturally understood by humans. We propose a training strategy for a 3D LiDAR semantic segmentation model that learns structural relationships between classes through abstraction. This is achieved by implicitly modeling these relationships using a learning rule for hierarchical multi-label classification (HMC). Our detailed analysis demonstrates that this training strategy not only improves the model's confidence calibration but also retains additional information useful for downstream tasks such as fusion, prediction, and planning.
\end{abstract}

\section{Introduction}\label{sec:intro}
Accurate 3D semantic segmentation is crucial for autonomous vehicles to achieve comprehensive scene understanding, supporting tasks like sensor fusion and prediction for detailed environmental modeling. However, these algorithms face challenges in ambiguous environments and noise. Recognizing and adapting to these limitations is essential. Low-confidence classifications, though potentially informative, may be disregarded during sensor fusion, resulting in critical data loss for subsequent tasks. This information is vital for planning algorithms to accurately identify and assess planning-relevant instances, such as vulnerable road users (VRUs). Failure to recognize VRUs as dynamic objects can lead to hazardous situations, particularly concerning VRU protection \cite{Wiki2023}.

\begin{figure}[t!]
	\centering
	\begin{subfigure}[b]{0.3\columnwidth}
		\centering
		\includegraphics[width=\textwidth]{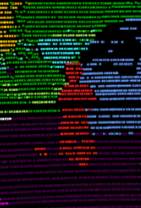}
	\end{subfigure}
	\hfill
	\begin{subfigure}[b]{0.3\columnwidth}
		\centering
		\includegraphics[width=\textwidth]{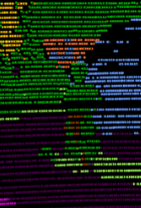}
	\end{subfigure}
	\hfill
	\begin{subfigure}[b]{0.3\columnwidth}
		\centering
		\includegraphics[width=\textwidth]{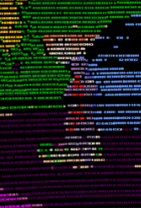}
	\end{subfigure}
	\\\scriptsize{\textd{sidewalk}, \textd{vegetation}, \textd{terrain}, \textd{car}, \textd{building}, \\ \textd{bicycle}, \textd{person}, \textd{trunk}, \textd{road}, \textd{dynamic}, \textd{static}, \any}
	\caption{The point cloud visualization of a nearby kid with a scooter (left: ground truth) from the SemanticKITTI \cite{Behley2019} dataset shows a stark difference between the models. A hierarchy-agnostic model (mid) misclassifies it as \class{vegetation}, whereas our hierarchy-aware model (right) correctly identifies it as \class{dynamic}.}
	\label{fig:teaser}
\end{figure}

In this study, we propose a method to provide confident and well-calibrated estimates of instance types at a point-wise level for downstream tasks. We emphasize the importance of high-confidence, detail-agnostic information -- such as identifying static or dynamic objects, VRUs, traffic signs, and drivable spaces. To achieve this, we employ a hierarchical multi-label classification (HMC) training strategy for 3D semantic segmentation models, considering structural similarities between semantically related classes. This approach explicitly models relationships to inform the model about alternative class representations and abstractions, as depicted in Fig. \ref{fig:teaser}. Providing high-level information, including abstract attributes like \class{dynamic}, mitigates the risk of overlooking critical factors in tasks such as vehicle trajectory planning.

We propose a learning rule enabling classification models to provide detailed classifications in clear situations while predicting superclasses in ambiguous circumstances. This approach enhances overall predictive confidence and preserves critical information for downstream tasks. Additionally, we introduce a method to extract implicit model uncertainties from hierarchical classifications. Our analysis quantitatively and qualitatively explores how semantic label definitions and inter-class structural relationships impact model training. We evaluate the model's confidence calibration and abstraction capabilities under a domain shift. Our goal is to optimize the balance between detail level and predictive confidence to maximize information utility in autonomous driving applications. Our contributions can be summarized as follows:
\begin{itemize}
\item Introduction a novel approach tailored for 3D semantic segmentation models, enhancing their ability to capture hierarchical relationships between classes.
\item Proposal of a technique to derive reliable confidence scores from hierarchical classifications, improving the model's certainty in complex environments.
\item Presentation of an evaluation metric to measure the HMC model's ability to generalize across semantic relationships.
\end{itemize}
\section{Related Literature}\label{sec:literature}
Extensive research efforts have focused on advancing semantic segmentation algorithms for point clouds, categorized by their underlying data representation. These methods include projection-based approaches, such as range view \cite{Miloto2019,Cortinhal2020} and bird's eye view \cite{Zhang2020}; point-based methods, such as unordered point clouds \cite{Qi2017} and point pillars \cite{Fei2021}; and voxel-based methods, including volumetric 3D convolutions \cite{Zhu2021} and sparse convolutions \cite{Tang2020,Xu2021}.

Semantic segmentation poses a unique challenge due to class imbalances. The widely used mean Intersection over Union (IoU) metric addresses this by ensuring equitable consideration of underrepresented classes. To mitigate imbalance in model classifications, common approaches include loss weighting to favor underrepresented classes \cite{Paszke2016,Miloto2019} or designing architectures capable of handling challenges posed by smaller instances \cite{Piewak2018,Xu2021,Chen2021}. 
However, these methods typically do not exploit structural similarities between classes to effectively alleviate class imbalance inherent in semantic segmentation tasks.

Moreover, in the context of autonomous driving, not all class confusions hold equal significance. The authors of \cite{Liu2020Uncertainty} and \cite{Zhou2023} delve into assessing the severity of class confusions in semantic segmentation. They respectively propose a method and a metric that evaluate the criticality of these confusions.

\subsection{Uncertainty Estimation in Classification Tasks} 
Recent advancements in uncertainty estimation have distinguished between two types: epistemic (model) uncertainty and aleatoric (data) uncertainty \cite{Arnez2020,Huellermeier2021}. Aleatoric uncertainty arises from data variability and can be learned during training, as demonstrated by methods like sampling from logits before applying softmax \cite{Kendall2017}. Addressing epistemic uncertainty remains challenging, with Monte-Carlo Dropout (MCD, \cite{Gal2016}) and deep ensembles (DE, \cite{Lakshminarayanan2017}) being established approaches. 

These techniques, though computationally intensive, improve softmax calibration by approximating posterior distributions across classes \cite{Guo2017}. Recent research has focused on deterministic methods for estimating epistemic uncertainty \cite{Postels2022,Mukhoti2023}. Unlike sampling-based methods, deterministic approaches avoid additional computational complexity but are constrained by the model's inherent limitations. However, these insights are often underutilized, contributing to the information loss highlighted in Section \ref{sec:intro}.

\subsection{Hierarchical Multi-Label Classification}
Hierarchical multi-label classification (HMC), in contrast to hierarchy-agnostic approaches, leverages structural relationships between class labels to enhance the performance and robustness of classification models \cite{Levatic2014,Zhang2017,Chen2021}. Instead of treating class labels as independent entities, HMC arranges them in a hierarchical tree structure. This approach allows data points to potentially belong to multiple classes simultaneously. This has been applied in biomedical applications and image and text classification \cite{Wehrmann2018,Xu2019,Giunchiglia2020,Romero2022}.

Some studies leverage the structural relationships between classes during training, focusing exclusively on leaf classes \cite{Li2022,Vaswani2022}. Others, such as \cite{deGraaff2022} and \cite{Ke2022}, harness the hierarchical semantic relationships to address complex tasks like novelty detection and unsupervised semantic segmentation. While these methods utilize learned relational connections to enhance classification performance, they often do not fully exploit these relationships for abstract class classification in uncertain scenarios. We aim to achieve that with our proposed hierarchical multi-label classification (HMC) approach.
\section{Methods}\label{sec:methods}
The following section outlines our proposed modifications designed to transform any architecture from a hierarchy-agnostic model to a hierarchy-aware model.

\subsection{Data and Label Hierarchy}
\begin{figure}[h]
	\centering
	\includegraphics[width=0.6\columnwidth]{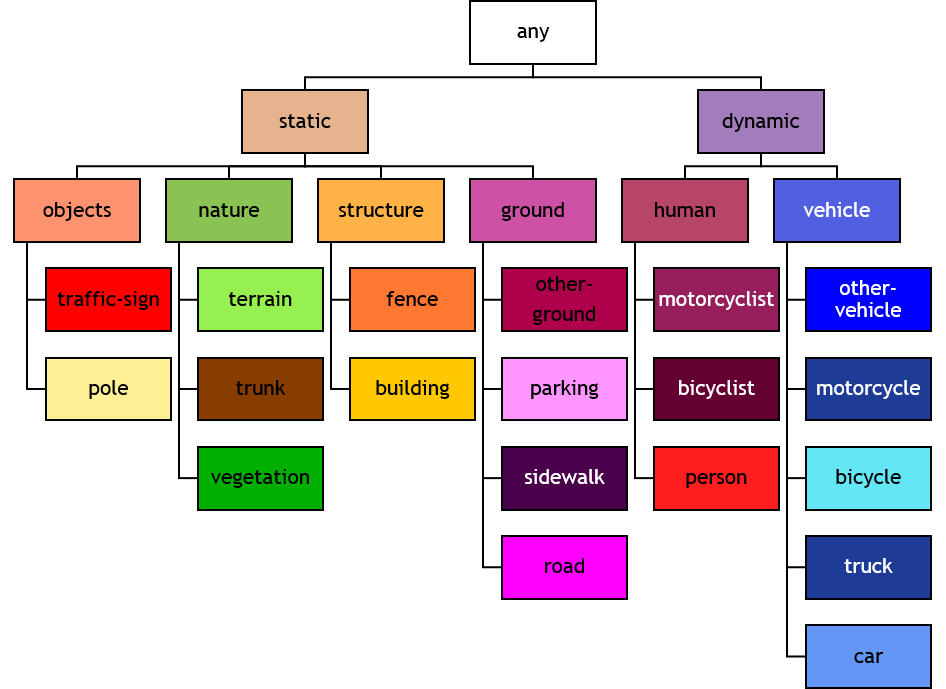}
	\caption{Label hierarchy for the SemanticKITTI dataset: the original labels are leaf nodes, meta and binary classes are added accordingly. The colors denote the label colors as used in the dataset (except for the \class{bicyclist} class, whose color is adjusted for improved visibility).}
	\label{fig:hierarchy}
\end{figure}

The label hierarchy serves to group semantically similar classes together. Our proposed label hierarchy for the SemanticKITTI dataset is illustrated in Fig. \ref{fig:hierarchy}. Each original label is associated with multiple superclasses $\mathcal{S}$ on a level $l$. The dataset's original ground truth labels $y_s$ represent the leaf nodes ($l=0$). Superclasses are structured as follows: the second level aligns with SemanticKITTI's suggestion ($l=1$), and the third level introduces binary classes, namely \textit{static} and \textit{dynamic} ($l=2$). The root node encompasses only the \textit{any} class ($l=3$). Therefore, the height of this label hierarchy tree is $h=4$.

The ground truth $\eta_c$ for HMC training is derived from this hierarchy. We define it such that more than one label can be correct, but never more than one per hierarchy level $l$. Therefore, the HMC learning rule incorporates multiple correct labels that are learned jointly. The entirety of all classes is denoted as $\lambda$, and $\lambda_{l}$ represents the set of classes at hierarchy level $l$. To prevent the model from exclusively predicting the \textit{any} class, we assign weights to the superclasses $\mathcal{S}(y_s)$ of each labeled class $y_s$ based on its hierarchy level $l$. Specifically, labels in the leaf nodes $\lambda_{0}$ are assigned the highest weights, while superclasses $\lambda_{1}$ receive lower weights, and so forth. We propose the following formula to weight the new ground truth:
\begin{equation}\label{eq:altered_gt}
	\eta_c = 
	\begin{cases}
		\frac{1 + l(c)}{h}, & \text{if } c \in y_s\cup\mathcal{S}(y_s)\\
		0, & \text{otherwise}
	\end{cases}
\end{equation}
The altered ground truth $\eta_c$ is computed from the original subclass $y_s$ and its superclasses $\mathcal{S}(y_s)$. For all other classes it is assigned 0. According to this ground truth definition, multiple labels assigned to a single measurement trace a path through the hierarchy, reflecting their structural similarity. This implicit relationship modeling enables the model to leverage these relationships for enhanced leaf-node classification as well as meaningful superclass classifications.

\subsection{Hierarchical Multi-Label Classification}\label{sub:models}
The HMC loss function incorporates the weighted label encoding as presented in Eq. \ref{eq:altered_gt}. We propose a learning rule that integrates the label hierarchy into the model training process:
\begin{equation}\label{eq:exploss}
    \mathcal{L}_{HCE} = -\sum_{c \in \lambda} e^{\eta_c} \log \hat{y}_c
\end{equation}
with $c$ being a given class in $\lambda$, and $\eta_c$ and $\hat{y}_c$ being the hierarchically altered ground truth and the model classifications for that class, respectively.

To benchmark our proposed model against hierarchy-agnostic models, we applied a heuristic hierarchy construction to baseline models using their confidence outputs. Predictions ascend to a higher hierarchy level when accompanied by low confidence. Thresholds are evenly distributed across the hierarchy, defined as $\delta = \frac{l}{h}$: predictions remain at level 0 above $0.75$, move to level 2 at $0.5$, level 1 at $0.25$, and below $0.25$ ascend to level 3. This equidistant spacing considers the model's calibration, ensuring that better-calibrated models reflect a consistent confidence structure.

\subsection{Hierarchical Classification and Confidence Estimation}\label{sub:ue}
Our objective with the proposed HMC training strategy is twofold: \textbf{(a)} Enhance classification quality by leveraging the hierarchical structure, where the model yields high confidences corresponding to the level of detail achievable based on the data. For that, high-level classification information is provided and can be confidently utilized by downstream tasks. \textbf{(b)} Capture uncertainty through the hierarchical predictions, enabling the model to express confidences specifically for classifications at the finest detail level (leaf classes). Here, the model delivers well-calibrated results specifically at the finest detail level (leaf classes), which is advantageous when superclass classifications are undesired. We propose reporting distinct confidence measures based on the HMC classifications to address this dual objective.

To achieve the first objective, we employ the softmax function over the entire hierarchy $\lambda$. This function transforms the model's output logits into class-specific probabilities:
\begin{equation}\label{eq:p_sm}
	p_\sigma = \frac{e^{x}}{\sum_{c=1}^{\lambda}e^{c}}
\end{equation}
with $x$ representing the logit vector, and $c$ denoting all classes in the hierarchy $\lambda$, we utilize the softmax function. We calculate $\arg\max(p_{\sigma}(\lambda))$ to ensure consistent high-confidence classifications across all levels of the label hierarchy.  

To capture model uncertainty based on the classification level in the hierarchy, we compute entropy over the softmax probabilities of the original leaf classes, denoted as $p^l = p_\sigma \in p_\sigma(\lambda_{0})$. When the model predicts a leaf class in straightforward scenarios, $\arg\max(p_{\sigma}(\lambda))$ corresponds to the lowest hierarchy level. In uncertain situations, where the model predicts a superclass, softmax values are more evenly distributed among the leaf classes of the predicted superclasses, resulting in higher entropy among the leaf nodes. Thus, predictive uncertainty is quantified by the entropy of these leaf classes:
\begin{equation}\label{eq:p_H}
    p_\mathcal{H} = - p^l \cdot \log(p^l)
\end{equation}
Since entropy $\mathcal{H}$ is not a probabilistic measure, we normalize it using the theoretical maximum entropy value $\log(n)$ with $n$ being the number of classes.

\begin{figure}[h]
	\centering
	\begin{subfigure}[b]{\columnwidth}
		\centering
		\includegraphics[width=\textwidth]{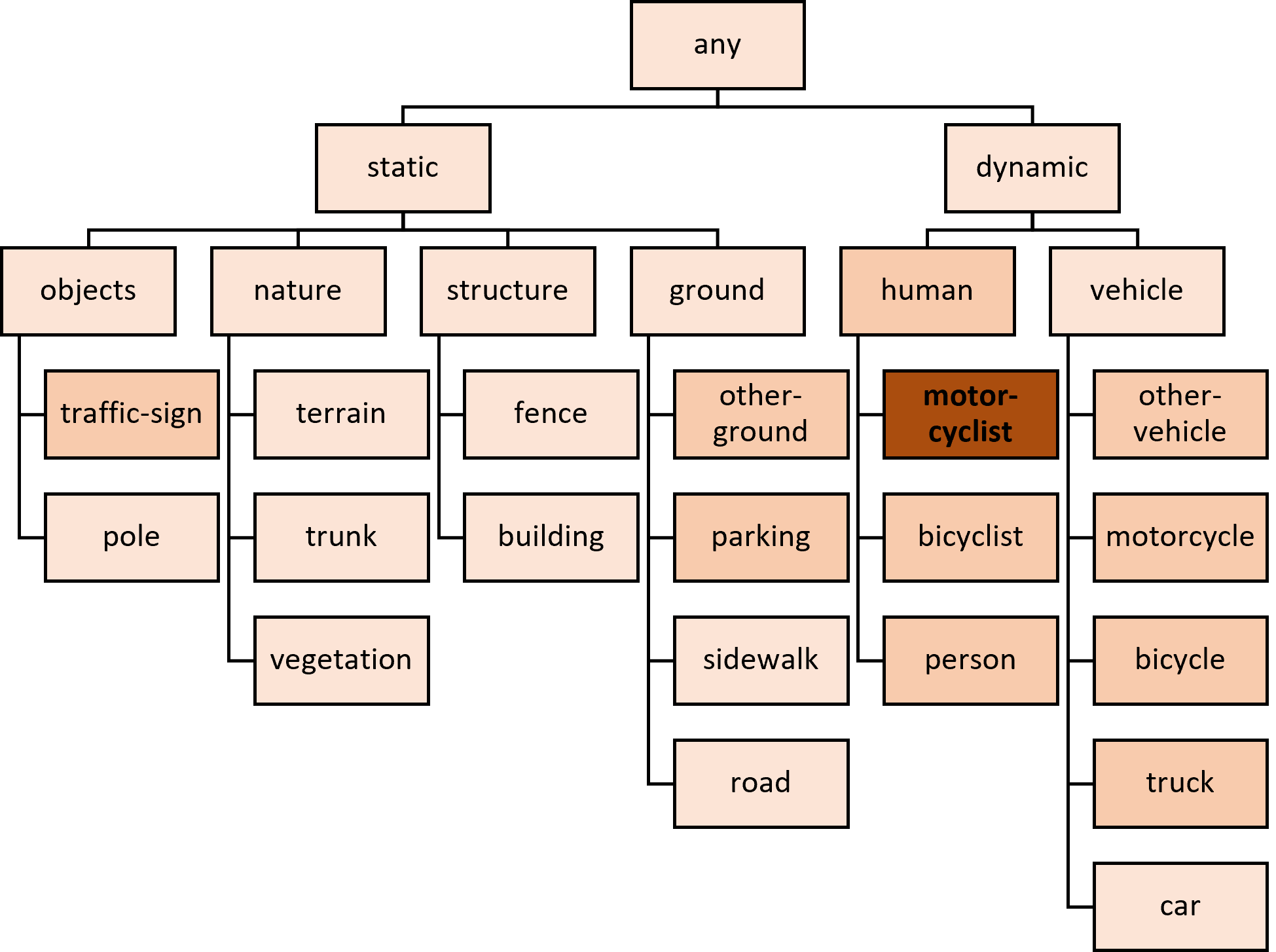}
		\caption{Averaged softmax probabilities for the class \class{motorcyclist}.}
		\label{subfig:probs_motorcyclist}
	\end{subfigure}
	\\
	\begin{subfigure}[b]{\columnwidth}
		\centering
		\includegraphics[width=\textwidth]{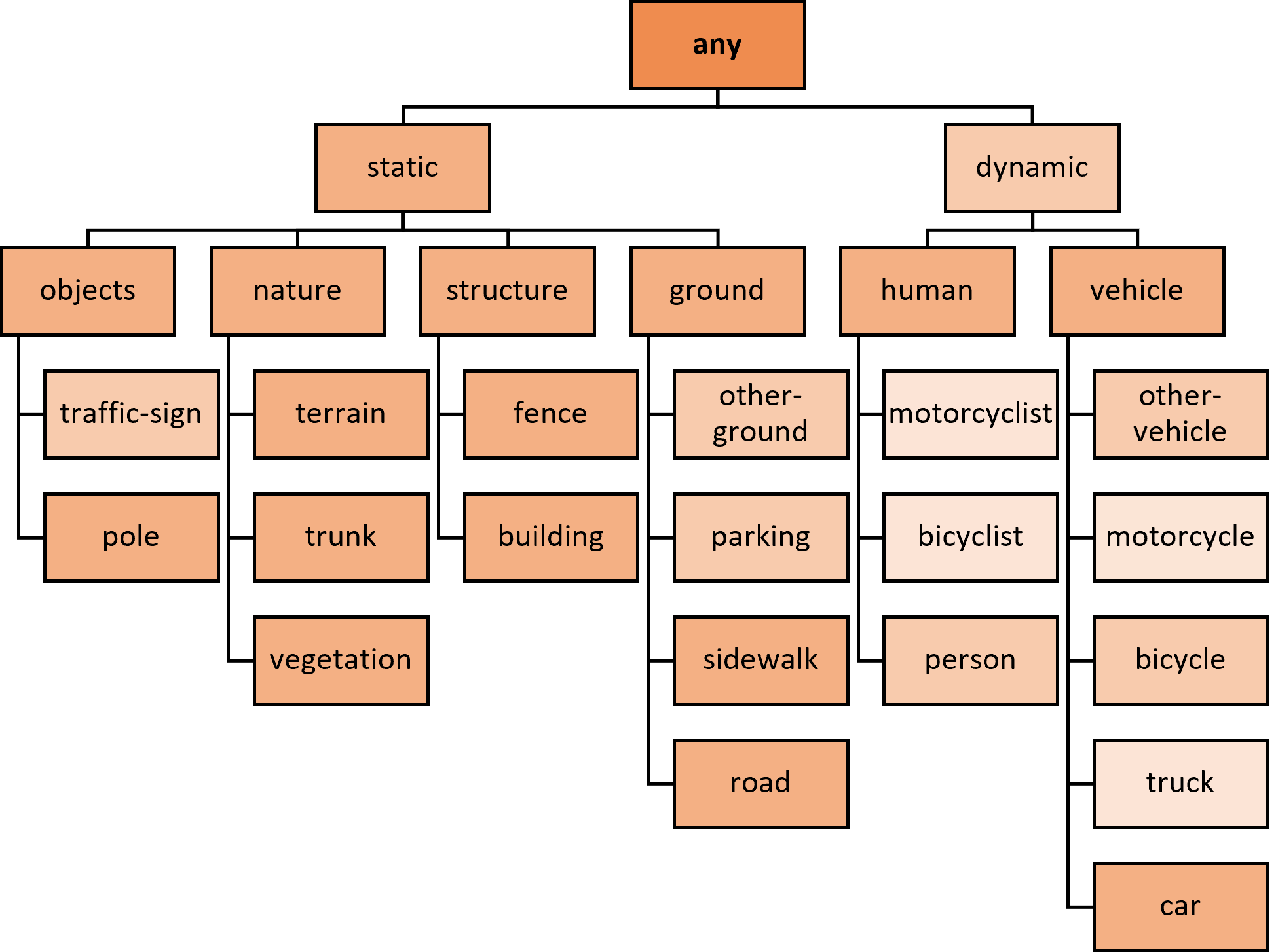}
		\caption{Averaged softmax probabilities for the class \class{any}.}
		\label{subfig:probs_any}
	\end{subfigure}
	\\\scriptsize{predictive probabilities: }\\ \includegraphics[width=0.75\columnwidth]{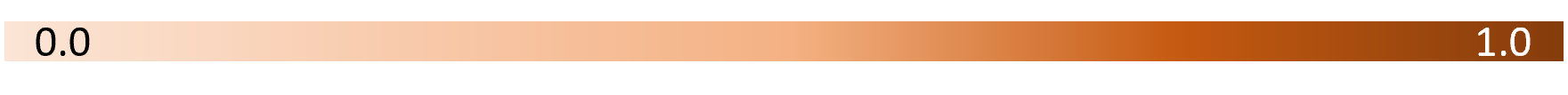}
	\caption{Softmax probabilities $\sigma(\lambda)$ for classifications of classes \class{motorcyclist} (\ref{subfig:probs_motorcyclist}) and \class{any} (\ref{subfig:probs_any}). The colorscale is given below.}
	\label{fig:probs}
\end{figure}

The relationship between these measures is demonstrated in Fig. \ref{fig:probs}, depicting the label hierarchy from Fig. \ref{fig:hierarchy}. Classes are color-coded based on their average predictive probabilities using $p_{\sigma}$. Instances where the model confidently predicts leaf classes yield low uncertainty (e.g., $\mathcal{H} \sim 0.3$ for \textit{motorcyclist}), whereas high uncertainty (e.g., $\mathcal{H} \sim 1.0$ for \textit{any}) arises when softmax probabilities are evenly distributed across leaf classes due to superclass predictions.

\subsection{Calibration Metrics for Semantic Segmentation}\label{sub:metrics}
Since the HMC model can predict abstract superclasses, evaluation metrics for semantic segmentation models need adjustment to ensure fair comparison with hierarchy-agnostic baseline models. The commonly used mIoU metric evaluates True Positives (\textit{TP}), False Positives (\textit{FP}), and False Negatives (\textit{FN}). The Critical Error Rate (CER, \cite{Zhou2023}) metric evaluates misclassifications within specific class categories. It identifies critical errors by assessing \textit{FP} and \textit{FN} within expected categories. To comprehensively evaluate hierarchical classification, we propose the hierarchical IoU (hIoU). During training, superclass classifications are treated as partially correct (cf. Eq. \ref{eq:exploss}). For evaluation, we extend the regular class-wise IoU by incorporating superclass predictions in a similar manner. By default, superclass predictions are considered incorrect, as they are not included in the actual ground truth. We introduce a modulation factor $\alpha \in [0.0..1.0]$ to adjust the weight of correct superclass classifications in the final metric value, referred to as partial True Positives (\textit{pTP}). Thus, at $\alpha = 0.0$, all superclass predictions are deemed incorrect, while at $alpha = 1.0$, superclasses are considered as correct as the original leaf classes. The $hIoU_s$ for subclass $s \in \lambda_{0}$ is defined as:
\begin{equation}
	hIoU_s = \frac{TP_s + pTP_s}{TP_s+FP_s+FN_s} = \frac{TP_s + \sum_{l=1}^{h-1} \alpha^l \cdot TS_s(l)}{TP_s+FP_s+FN_s}
\end{equation} 
where \textit{TS}$(l)$ denotes the True Superclasses at each level $\lambda_{1..3}$. All superclass predictions \textit{FS}, whether true or false, are included in \textit{FNs}. The overall hIoU is the mean over all original leaf classes: $hIoU = \frac{1}{\lambda_0}\sum_{0}^{\lambda_0}hIoU_s$. 
\begin{figure*}[]
	\centering
	\begin{subfigure}[b]{0.13\textwidth}
		\centering
		\includegraphics[width=\textwidth]{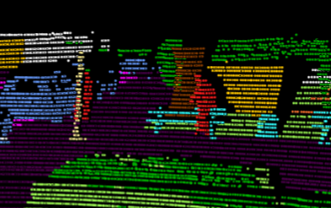}
	\end{subfigure}
	\hfill
	\begin{subfigure}[b]{0.13\textwidth}
		\centering
		\includegraphics[width=\textwidth]{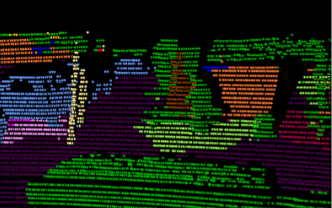}
	\end{subfigure}
	\hfill
	\begin{subfigure}[b]{0.13\textwidth}
		\centering
		\includegraphics[width=\textwidth]{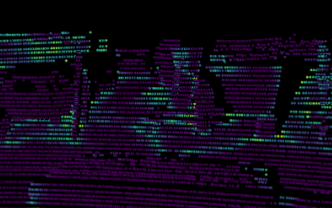}
	\end{subfigure}
	\hfill
	\begin{subfigure}[b]{0.13\textwidth}
		\centering
		\includegraphics[width=\textwidth]{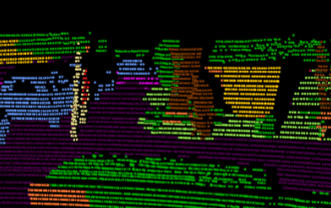}
	\end{subfigure}
	\hfill
	\begin{subfigure}[b]{0.13\textwidth}
		\centering
		\includegraphics[width=\textwidth]{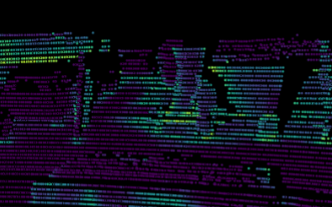}
	\end{subfigure}
	\hfill
	\begin{subfigure}[b]{0.13\textwidth}
		\centering
		\includegraphics[width=\textwidth]{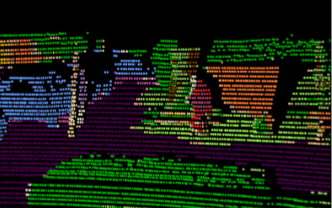}
	\end{subfigure}
	\hfill
	\begin{subfigure}[b]{0.13\textwidth}
		\centering
		\includegraphics[width=\textwidth]{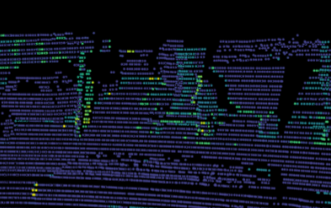}
	\end{subfigure}
	\hfill 
	\begin{subfigure}[b]{0.13\textwidth}
		\centering
		\includegraphics[width=\textwidth]{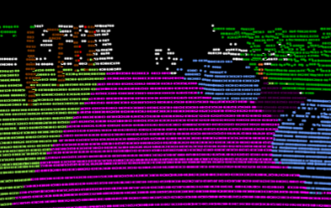}
	\end{subfigure}
	\hfill
	\begin{subfigure}[b]{0.13\textwidth}
		\centering
		\includegraphics[width=\textwidth]{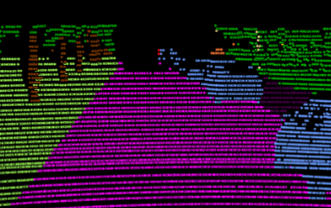}
	\end{subfigure}
	\hfill
	\begin{subfigure}[b]{0.13\textwidth}
		\centering
		\includegraphics[width=\textwidth]{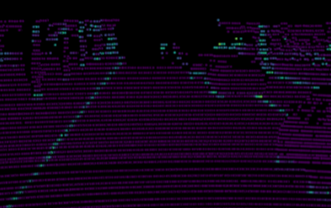}
	\end{subfigure}
	\hfill
	\begin{subfigure}[b]{0.13\textwidth}
		\centering
		\includegraphics[width=\textwidth]{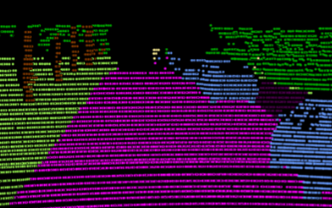}
	\end{subfigure}
	\hfill
	\begin{subfigure}[b]{0.13\textwidth}
		\centering
		\includegraphics[width=\textwidth]{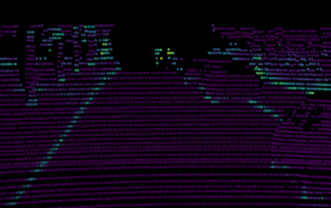}
	\end{subfigure}
	\hfill
	\begin{subfigure}[b]{0.13\textwidth}
		\centering
		\includegraphics[width=\textwidth]{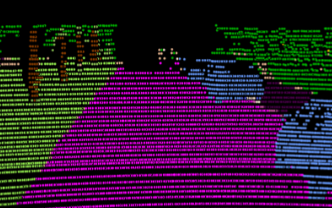}
	\end{subfigure}
	\hfill
	\begin{subfigure}[b]{0.13\textwidth}
		\centering
		\includegraphics[width=\textwidth]{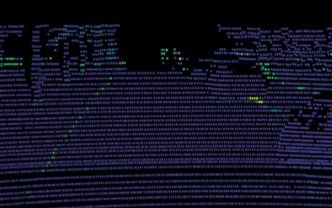}
	\end{subfigure}
	\hfill 
	\begin{subfigure}[b]{0.13\textwidth}
		\centering
		\includegraphics[width=\textwidth]{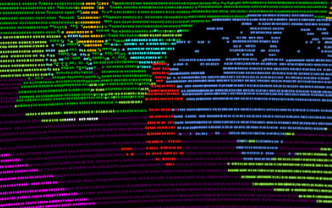}
	\end{subfigure}
	\hfill
	\begin{subfigure}[b]{0.13\textwidth}
		\centering
		\includegraphics[width=\textwidth]{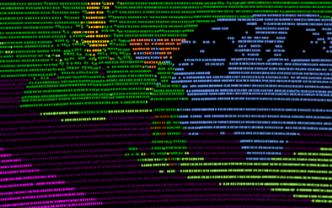}
	\end{subfigure}
	\hfill
	\begin{subfigure}[b]{0.13\textwidth}
		\centering
		\includegraphics[width=\textwidth]{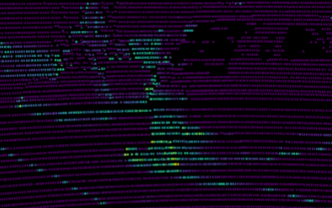}
	\end{subfigure}
	\hfill
	\begin{subfigure}[b]{0.13\textwidth}
		\centering
		\includegraphics[width=\textwidth]{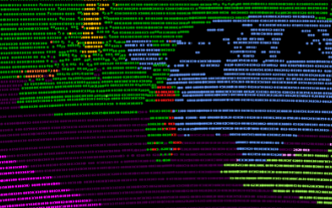}
	\end{subfigure}
	\hfill
	\begin{subfigure}[b]{0.13\textwidth}
		\centering
		\includegraphics[width=\textwidth]{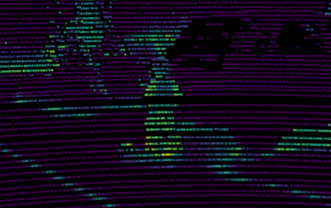}
	\end{subfigure}
	\hfill
	\begin{subfigure}[b]{0.13\textwidth}
		\centering
		\includegraphics[width=\textwidth]{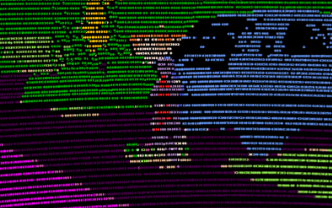}
	\end{subfigure}
	\hfill
	\begin{subfigure}[b]{0.13\textwidth}
		\centering
		\includegraphics[width=\textwidth]{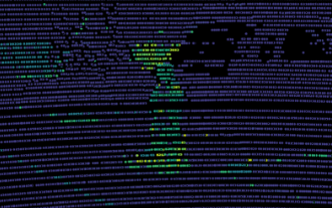}
	\end{subfigure}
	\hfill 
	\begin{subfigure}[b]{0.13\textwidth}
		\centering
		\includegraphics[width=\textwidth]{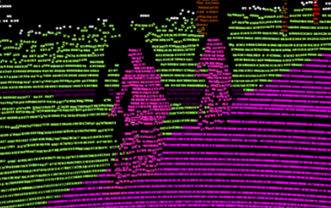}
	\end{subfigure}
	\hfill
	\begin{subfigure}[b]{0.13\textwidth}
		\centering
		\includegraphics[width=\textwidth]{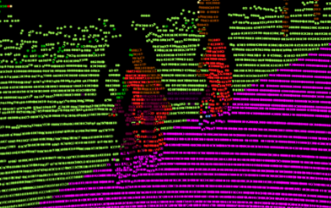}
	\end{subfigure}
	\hfill
	\begin{subfigure}[b]{0.13\textwidth}
		\centering
		\includegraphics[width=\textwidth]{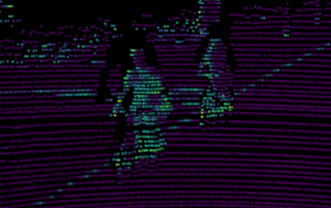}
	\end{subfigure}
	\hfill
	\begin{subfigure}[b]{0.13\textwidth}
		\centering
		\includegraphics[width=\textwidth]{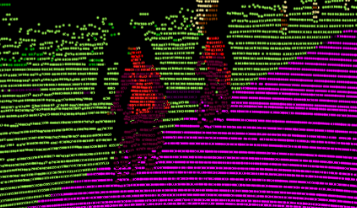}
	\end{subfigure}
	\hfill
	\begin{subfigure}[b]{0.13\textwidth}
		\centering
		\includegraphics[width=\textwidth]{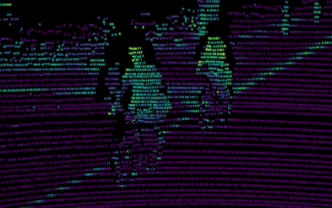}
	\end{subfigure}
	\hfill
	\begin{subfigure}[b]{0.13\textwidth}
		\centering
		\includegraphics[width=\textwidth]{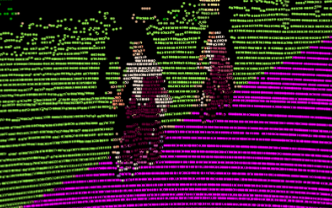}
	\end{subfigure}
	\hfill
	\begin{subfigure}[b]{0.13\textwidth}
		\centering
		\includegraphics[width=\textwidth]{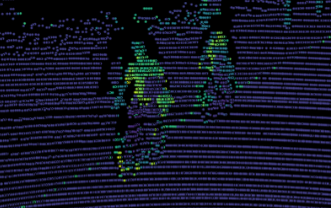}
	\end{subfigure}
	\hfill 
	\begin{subfigure}[b]{0.13\textwidth}
		\centering
		\includegraphics[width=\textwidth]{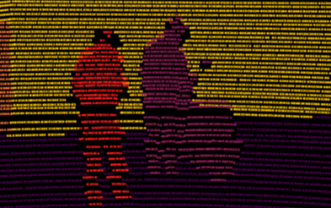}
	\end{subfigure}
	\hfill
	\begin{subfigure}[b]{0.13\textwidth}
		\centering
		\includegraphics[width=\textwidth]{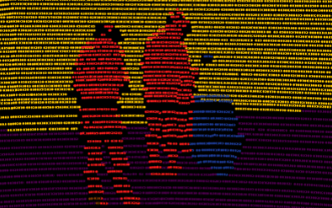}
	\end{subfigure}
	\hfill
	\begin{subfigure}[b]{0.13\textwidth}
		\centering
		\includegraphics[width=\textwidth]{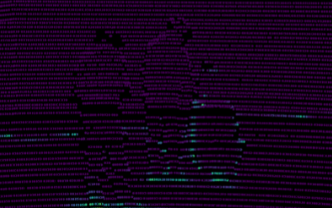}
	\end{subfigure}
	\hfill
	\begin{subfigure}[b]{0.13\textwidth}
		\centering
		\includegraphics[width=\textwidth]{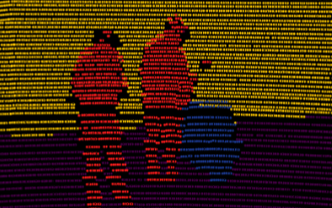}
	\end{subfigure}
	\hfill
	\begin{subfigure}[b]{0.13\textwidth}
		\centering
		\includegraphics[width=\textwidth]{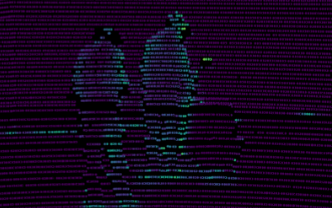}
	\end{subfigure}
	\hfill
	\begin{subfigure}[b]{0.13\textwidth}
		\centering
		\includegraphics[width=\textwidth]{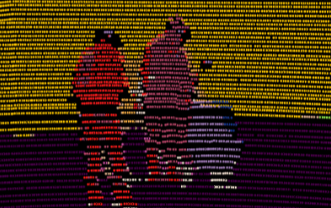}
	\end{subfigure}
	\hfill
	\begin{subfigure}[b]{0.13\textwidth}
		\centering
		\includegraphics[width=\textwidth]{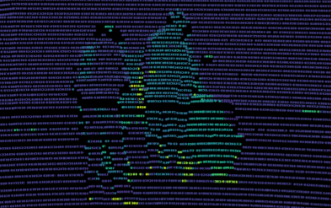}
	\end{subfigure}
	\hfill 
	\begin{subfigure}[b]{0.13\textwidth}
		\centering
		\includegraphics[width=\textwidth]{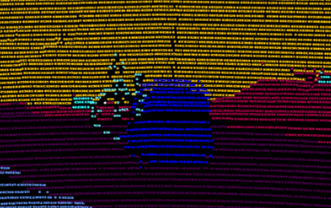}
	\end{subfigure}
	\hfill
	\begin{subfigure}[b]{0.13\textwidth}
		\centering
		\includegraphics[width=\textwidth]{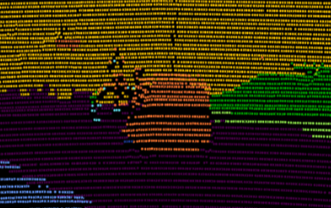}
	\end{subfigure}
	\hfill
	\begin{subfigure}[b]{0.13\textwidth}
		\centering
		\includegraphics[width=\textwidth]{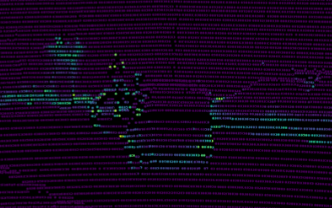}
	\end{subfigure}
	\hfill
	\begin{subfigure}[b]{0.13\textwidth}
		\centering
		\includegraphics[width=\textwidth]{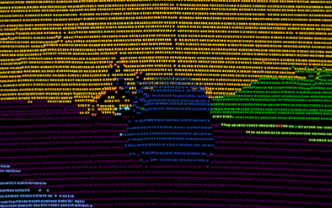}
	\end{subfigure}
	\hfill
	\begin{subfigure}[b]{0.13\textwidth}
		\centering
		\includegraphics[width=\textwidth]{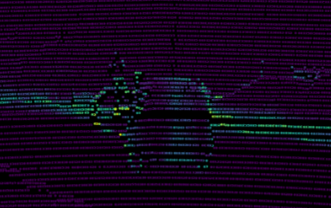}
	\end{subfigure}
	\hfill
	\begin{subfigure}[b]{0.13\textwidth}
		\centering
		\includegraphics[width=\textwidth]{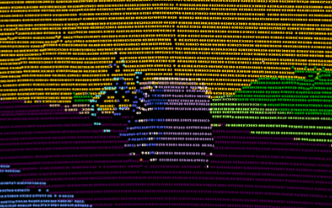}
	\end{subfigure}
	\hfill
	\begin{subfigure}[b]{0.13\textwidth}
		\centering
		\includegraphics[width=\textwidth]{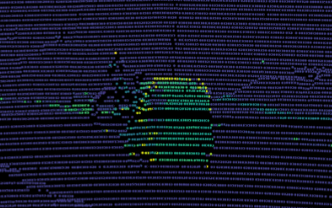}
	\end{subfigure}
	\hfill 
	\begin{subfigure}[b]{0.13\textwidth}
		\centering
		\includegraphics[width=\textwidth]{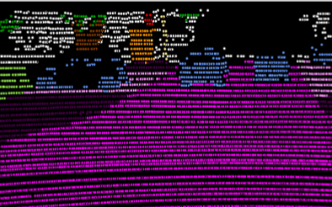}
		\caption{Original labels}
		\label{subfig:label}
	\end{subfigure}
	\hfill
	\begin{subfigure}[b]{0.13\textwidth}
		\centering
		\includegraphics[width=\textwidth]{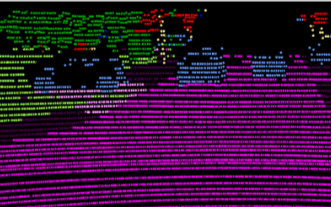}
		\caption{Vanilla pred.}
		\label{subfig:flatpred}
	\end{subfigure}
	\hfill
	\begin{subfigure}[b]{0.13\textwidth}
		\centering
		\includegraphics[width=\textwidth]{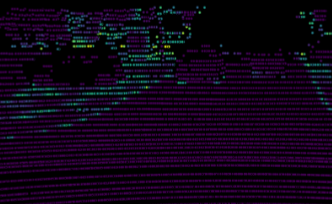}
		\caption{Vanilla conf.}
		\label{subfig:flatconf}
	\end{subfigure}
	\hfill
	\begin{subfigure}[b]{0.13\textwidth}
		\centering
		\includegraphics[width=\textwidth]{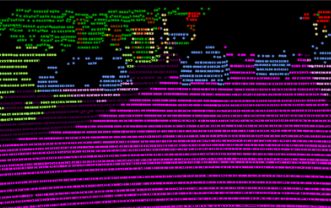}
		\caption{MCD pred.}
		\label{subfig:mcdpred}
	\end{subfigure}
	\hfill
	\begin{subfigure}[b]{0.13\textwidth}
		\centering
		\includegraphics[width=\textwidth]{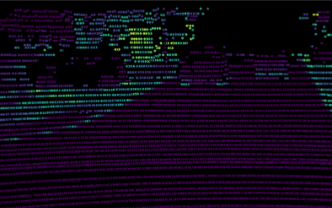}
		\caption{MCD conf.}
		\label{subfig:mcdconf}
	\end{subfigure}
	\hfill
	\begin{subfigure}[b]{0.13\textwidth}
		\centering
		\includegraphics[width=\textwidth]{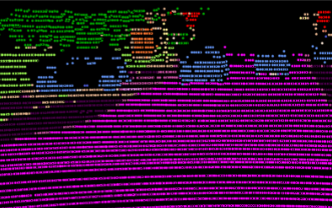}
		\caption{HMC pred.}
		\label{subfig:hmcpred}
	\end{subfigure}
	\hfill
	\begin{subfigure}[b]{0.13\textwidth}
		\centering
		\includegraphics[width=\textwidth]{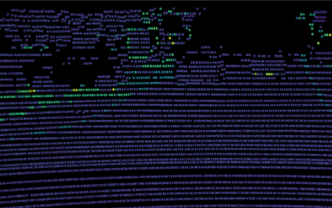}
		\caption{HMC conf.}
		\label{subfig:hmcconf}
	\end{subfigure}
	\\\scriptsize{predictive confidences: \\ \includegraphics[width=\columnwidth]{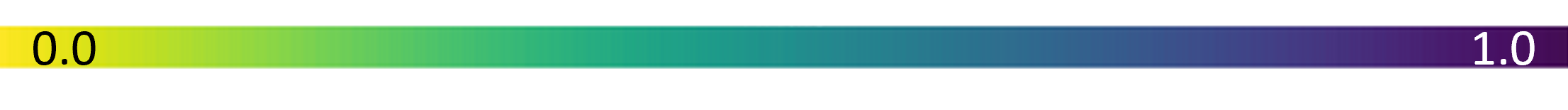}} 
	\caption{Qualitative samples from the SemanticKITTI \cite{Behley2019} dataset, showing the labels \ref{subfig:label} and the predictions and confidences of the vanilla \cite{Tang2020} (\ref{subfig:flatpred} \& \ref{subfig:flatconf}), MCD \cite{Cortinhal2020} (\ref{subfig:mcdpred} \& \ref{subfig:mcdconf}) and HMC model (\ref{subfig:hmcpred} \& \ref{subfig:hmcconf}, \pent is used as confidence measure). The semantic class colors are depicted as in \ref{fig:hierarchy}, the confidence colorscale is given below.}
	\label{fig:qualitative}
\end{figure*}
\section{Experimental Results}\label{sec:experiments}
In the following section, we present experimental results demonstrating the HMC model's proficiency in learning class abstractions and conveying uncertainty through superclass classifications. For all models, including the HMC, we evaluate $\arg\max(p_{\sigma})$ for classification and $p_{H}$ for confidence performances.

\textbf{Metrics.} In addition to quantitative analyses using metrics detailed in Section \ref{sub:metrics} (mIoU, hIoU, and CER), we evaluate the model's calibration. Calibration metrics, such as Expected Calibration Error (ECE \cite{Guo2017}), Area Under the Sparsification Error Curve (AUSE \cite{Ilg2018}), and uncertainty-aware mIoU (uIoU \cite{Sakaridis2022}), assess the alignment between predicted confidences and actual performance. ECE quantifies accuracy through binning, while AUSE progressively filters low-confidence predictions to evaluate model performance on more confident outputs, adaptable to metrics like Brier Score \cite{Gustafsson2020} or semantic segmentation mIoU \cite{Dreissig2023}. uIoU extends mIoU by considering invalid masks, distinguishing True Invalids (\textit{TI}) and False Invalids (\textit{FI}) based on a confidence threshold $\theta$, with the final uIoU computed by averaging over thresholds $\theta \in [0.0, 1.0]$. 

\textbf{Training Details.} All trainings were conducted on the training sequences of SemanticKITTI dataset \cite{Behley2019} on the voxelized point clouds. All models are trained for maximum $e=100$ epochs with a batch size of $bs=8$. The early stopping condition is set to a $\delta = 0.001$ on the validation mIoU for 10 epochs. As optimizer we chose Adam \cite{Kingma2014} with an initial learning rate of $10^{-4}$ and a cosine scheduler. 

\textbf{Baseline Models.} To benchmark our proposed model, we compared it against established uncertainty estimation methods. In addition to a plain cross-entropy loss-trained model (vanilla), we evaluated performance using sampling-based approaches with varying sample numbers to illustrate performance dependency. We selected logit sampling \cite{Kendall2017} with 10 and 15 samples (logit10, logit15) and Monte Carlo Dropout \cite{Gal2016} with 10 and 15 samples (MCD10, MCD15). Given that sampling-based methods require more computational power and inference time, we also included a deterministic baseline, Evidential Deep Learning (EDL) \cite{Sensoy2018}. Furthermore, we compared our HMC model with another hierarchy-based method (TM) \cite{Li2022}. The results presented here use the SPVCNN backbone \cite{Tang2020}; additional results on other 3D semantic segmentation backbones are provided in the supplemental material.

\subsection{Predictive Performance} 
To gain initial insights into the HMC model's performance compared to baseline models, we evaluate metrics such as mIoU, CER, and hIoU@$\alpha=1.0$ (hIoU1.0), along with relative runtime. The results, presented in Table \ref{tab:semKITTI_perf}, include relative performance gains and losses of our method compared to the best-performing baseline.

\begin{table}[h]
	\centering
	\begin{tabular}{@{}lrrrr@{}}
	\toprule
	method & mIoU $\uparrow$ & CER $\downarrow$ & hIoU1.0 $\uparrow$ & runtime $\downarrow$ \\ \midrule
	vanilla & 55.47 & 18.46 & 58.78 & +0.0\% \\
	logits10 \cite{Kendall2017} & 56.32 & 19.14 & 59.60 & +2.53\% \\
	logits15 \cite{Kendall2017} & 55.99 & 18.83 & 59.25 & +2.53\% \\
	MCD10 \cite{Gal2016} & 56.72 & 18.34 & 62.57 & +68.35\% \\
	MCD15 \cite{Gal2016} & 56.11 & 18.25 & 63.32 & +373.42\% \\
	EDL \cite{Sensoy2018} & 53.15 & 20.25 & 54.20 & +8.86\% \\
	TM \cite{Li2022} & 57.41 & 17.13 & 62.05 & \textbf{+0.0\%} \\
	HMC (ours) & \textbf{58.04} & \textbf{16.62} & \textbf{68.69} & +6.33\% \\ \bottomrule
	\end{tabular}
	\caption{Segmentation performance of the HMC and baseline models on the SemanticKITTI \cite{Behley2019} validation set. The best performances are marked in bold.}
	\label{tab:semKITTI_perf}
	\end{table}
The HMC model significantly outperforms all baselines, with only a 6.33\% increase in runtime. Runtime is calculated as the average time for one complete forward pass across the validation set, with the MCD15 model having a notably higher runtime due to increased memory demands.

The mIoU is calculated from leaf-only predictions in hierarchy-aware models ($p_\sigma(\lambda_0)$). Complementary, the model can predict superclass classifications by placing the softmax over the full tree hierarchy. An exemplary confusion matrix is given in the supplemental material. To understand the role of superclass predictions, we evaluate the hIoU metric at different $\alpha$ values in Table \ref{tab:hIoU}. Hierarchy-agnostic models are evaluated with a heuristic hierarchical rule as presented in Section \ref{sub:models}.

\begin{table*}[h]
	\centering
	\begin{tabular}{@{}lrrrrrrrrrrr@{}}
	\toprule
	\multirow{2}{*}{method} & \multicolumn{11}{c}{hIoU@$\alpha$ $\uparrow$} \\  
	 & 0.0 & 0.1 & 0.2 & 0.3 & 0.4 & 0.5 & 0.6 & 0.7 & 0.8 & 0.9 & 1.0 \\ \midrule
	vanilla & 54.79 & 55.16 & 55.54 & 55.92 & 56.31 & 56.71 & 57.11 & 57.52 & 57.93 & 58.35 & 58.78 \\
	logits10 \cite{Kendall2017} & \textbf{55.59} & \textbf{55.96} & 55.96 & 56.35 & 56.74 & 57.14 & 57.55 & 57.97 & 58.39 & 58.82 & 59.25 \\
	logits15 \cite{Kendall2017} & 55.21 & 55.58 & 56.34 & 56.72 & 57.11 & 57.51 & 57.91 & 58.32 & 58.74 & 59.17 & 59.60 \\
	MCD10 \cite{Gal2016} & 53.93 & 54.71 & 55.50 & 56.32 & 57.15 & 58.00 & 58.88 & 59.77 & 60.68 & 61.62 & 62.57 \\
	MCD15 \cite{Gal2016} & 54.61 & 55.42 & 56.24 & 57.08 & 57.94 & 58.82 & 59.72 & 60.64 & 61.58 & 62.54 & 63.52 \\
	EDL \cite{Sensoy2018} & 53.36 & 53.44 & 53.52 & 53.61 & 53.69 & 53.77 & 53.86 & 53.94 & 54.03 & 54.11 & 54.20 \\
	TM \cite{Li2022} & 55.28 & 55.82 & \textbf{56.39} & 56.99 & 57.63 & 58.29 & 58.98 & 59.70 & 60.45 & 61.24 & 62.05 \\
	HMC (ours) & 53.43 & 54.64 & 55.89 & \textbf{57.21} & \textbf{58.59} & \textbf{60.05} & \textbf{61.59} & \textbf{63.21} & \textbf{64.93} & \textbf{66.76} & \textbf{68.69}\\ \bottomrule
	\end{tabular}
	\caption{hIoU values for HMC and baseline models at different $\alpha$ values. The best performances per level are marked in bold.}
	\label{tab:hIoU}
	\end{table*}
Starting with a lower mIoU at $\alpha=0.0$, the HMC model quickly surpasses baseline models at an $\alpha$ as low as 0.3. This suggests that the HMC model effectively predicts superclasses in uncertain situations. Sampling-based approaches with more samples also outperform those with fewer samples at similar levels, likely due to improved calibration leading to increased number of superclass predictions, which are considered \textit{FP}s at lower $\alpha$ values. We propose that the $\alpha$ parameter can be tuned according to specific requirements for autonomous vehicles. For example, one requirement could be ensuring that VRUs are classified as dynamic objects in at least 90\% of cases. 

As discussed in \ref{sec:intro}, reliably reporting high-level information is crucial for fusion, prediction, and planning, particularly in autonomous driving where VRU protection is paramount. VRU classes are often underrepresented and exhibit high variance in semantic segmentation training data, making them difficult to learn. To illustrate the benefits of the HMC training strategy in safety-critical scenarios, Fig. \ref{fig:qualitative} presents examples featuring various VRU scenarios:
\begin{enumerate}
	\item The HMC model accurately detects a \class{human} missed by baseline models.
	\item Early detection of a distant bicyclist classified as \class{dynamic} by the HMC model, missed by baseline models.
	\item Detection of a nearby child, similar to previous VRU examples.
	\item Addressing challenges in classifying bicyclists, with the HMC model expressing uncertainty through superclass predictions.
	\item Similar difficulty in classifying a \class{motorcyclist}.
	\item Correct classification of a bicycle trailer (\class{other-vehicle}) as \class{dynamic} by the HMC model, with baseline models misclassifying it.
	\item Baseline models struggle to classify the transition from road to parking area accurately, crucial for planning drivable space.
\end{enumerate}

\subsection{Calibration and Out of Distribution Robustness}
To evaluate the HMC model's calibration, we use the entropy-based confidence scoring rule Eq. \ref{eq:p_H} and compare it with the entropy over the softmax for all baseline models. Calibration results are detailed in Table \ref{tab:calibration}, with relative performance gains and losses of the HMC model shown as percentages compared to the best baseline model.

\begin{table}[h]
	\centering
	\begin{tabular}{@{}lrrrr@{}}
	\toprule
	method & ECE $\downarrow$ & AUSE\textsubscript{BS} $\downarrow$ & AUSE\textsubscript{mIoU} $\downarrow$ & uIoU $\uparrow$ \\ \midrule
	vanilla & 17.24 & 1.93 & 12.82 & 84.56 \\
	logits10 \cite{Kendall2017} & 18.24 & 2.01 & 12.92 & 84.12 \\
	logits15 \cite{Kendall2017} & 16.55 & 2.07 & 13.05 & 84.52 \\
	MCD10 \cite{Gal2016} & 09.54 & 1.78 & 11.82 & 84.83 \\
	MCD15 \cite{Gal2016} & 10.68 & 1.85 & 12.58 & \textbf{85.66} \\
	EDL \cite{Sensoy2018} & 06.34 & 4.62 & 15.40 & 84.12 \\
	TM \cite{Li2022} & 09.34 & 2.14 & \textbf{9.33} & 84.55 \\
	HMC (ours) & \textbf{05.72} & \textbf{1.60} & 09.71 & 85.41 \\ \bottomrule
	\end{tabular}
	\caption{Calibration performances using entropy-based calibration measures. Best performances are marked in bold.}
	\label{tab:calibration}
	\end{table} 
The results underscore the challenging nature of assessing the calibration of point-wise classification models. The metrics do not indicate a single best model, as each metric evaluates different aspects of calibration. For instance, ECE, AUSE\textsubscript{BS}, and AUSE\textsubscript{mIoU} partially depend on binning and thresholding strategies, while uIoU is heavily influenced by base segmentation performance. Despite minor deviations, our HMC model demonstrates stable performance across all metrics and outperforms most baseline models. Interestingly, the hierarchy-aware TM model exhibits similar capabilities, likely due to its tree-structure training strategy. This approach, which considers the minimum confidence along the path from root to leaf nodes, results in better-calibrated softmax, and thus entropy, values.

To further stress-test the generalization abilities of the HMC trained models, we set up an anomaly detection setting by masking the \class{motorcyclist} class during the training. We expect the model to learn similarities to other VRU classes like \class{person} or \class{bicyclist}, based on appearance and position. 

\begin{table}[]
	\centering
	\begin{tabular}{@{}lrrrrrr@{}}
		\toprule
		mask & model & CER & \begin{tabular}[c]{@{}l@{}}IoU\\ (hIoU@\\ $\alpha$=0.0)\end{tabular} & \begin{tabular}[c]{@{}l@{}}hIoU@\\ $\alpha$=0.5\end{tabular} & \begin{tabular}[c]{@{}l@{}}hIoU@\\ $\alpha$=0.7\end{tabular} & \begin{tabular}[c]{@{}l@{}}hIoU@\\ $\alpha$=1.0\end{tabular} \\ \midrule
		\multirow{2}{*}{no} & vanilla & 20.80 & 0.00 & 2.76 & 3.93 & 5.78 \\
		& HMC & \textbf{19.83} & 0.00 & 4.90 & 8.00 & 14.06 \\ \midrule
		\multirow{2}{*}{yes} & vanilla & 88.88 & 0.00 & 5.23 & 7.50 & 11.11 \\
		& HMC & \textbf{74.36} & 0.00 & \bf{8.20} & \bf{13.92} & \bf{25.64} \\ \bottomrule
	\end{tabular}
	\caption{(h)IoU values in \% for the class \class{motorcyclist} scores (with and without being masked during training) for the baseline models and the HMC model at different $\alpha$s. The best performances are marked in bold.} \label{tab:motorcyclist_mask}
\end{table}

Tab. \ref{tab:motorcyclist_mask} shows the predictive performance for the class \class{motorcyclist} with and without being masked in the training dataset. Interestingly, the HMC model achieves an higher hIoU performance at higher $\alpha$ levels with the masked samples as with the learned samples. That means, that even for the hierarchy-agnostic baseline models but especially for the HMC model few, ambiguous samples of an underrepresented class are more harmful than none at all. Conclusively, the HMC is able to learn relationships between the classes, even without explicit labeling, probably due to the similarity in the feature space. This is an interesting insight for unsupervised novelty-detection and domain adaptation.

To examine the HMC model's generalization abilities on scene level, we expose it to a domain shift. All models are trained on the SemanticKITTI dataset \cite{Behley2019} and evaluated on the adverse weather SemanticSTF dataset \cite{Xiao2023}. Results are presented in Table \ref{tab:semSTF}, covering both segmentation and calibration performances.
\begin{table*}[h]
	\centering
	\begin{tabular}{@{}lrrrrrrr@{}}
	\toprule
	method & mIoU $\uparrow$ & CER $\downarrow$ & hIoU1.0 $\uparrow$ & ECE $\downarrow$ & AUSE\textsubscript{BS} $\downarrow$ & AUSE\textsubscript{mIoU} $\downarrow$ & uIoU $\uparrow$ \\ \midrule
	vanilla & 19.43 & 59.18 & 29.30 & 34.14 & 19.03 & 13.16 & 40.58 \\
	logits10 \cite{Kendall2017} & 22.66 & 54.19 & 33.38 & 34.20 & 18.98 & 15.31 & 42.42 \\
	logits15 \cite{Kendall2017} & 20.43 & 59.77 & 30.83 & 34.61 & 17.46 & 14.56 & 40.61 \\
	MCD10 \cite{Gal2016} & 20.46 & 59.05 & 36.67 & \textbf{27.74} & 14.69 & 15.03 & 40.39 \\
	MCD15 \cite{Gal2016} & 20.44 & 58.78 & 38.06 & 28.55 & 16.83 & 16.39 & 42.73 \\
	EDL \cite{Sensoy2018} & 17.34 & 58.58 & 19.13 & 45.68 & 21.62 & 15.59 & 36.97 \\
	TM \cite{Li2022} & 21.49 & 56.76 & 31.47 & 34.81 & 13.50 & 10.84 & 42.88 \\
	HMC (ours) & \textbf{22.77} & \textbf{54.58} & \textbf{39.75} & 36.06 & \textbf{10.32} & \textbf{10.28} & \textbf{43.73} \\ \bottomrule
	\end{tabular}
	\caption{Domain generalization results on the SemanticSTF \cite{Xiao2023} dataset, trained on SemanticKITTI data \cite{Behley2019}. Best performances are marked in bold.}
	\label{tab:semSTF}
	\end{table*}
Our HMC model demonstrates strong performance across most metrics, except for ECE. This underscores the potential of hierarchy-aware semantic classification models in open-world settings, effectively handling ambiguous and previously unseen scenes and objects. The challenging nature of the domain shift dataset, with adverse weather and unfamiliar surroundings, further emphasizes the robustness and adaptability of the HMC model.
\section{Conclusion}\label{sec:conclusion}
This study employed a hierarchical multi-label classification approach for 3D semantic segmentation of LiDAR point clouds in autonomous driving. By organizing class labels hierarchically, our model learns abstract class representations using inherent structural information and defaults to superclass classifications in ambiguous scenarios. This strategy not only reduces overall uncertainty but also preserves valuable information.

Through detailed analysis we demonstrated the efficacy of our approach in identifying safety-critical classes that are challenging to learn from 3D LiDAR data representations. We argue that confidently predicting high-level details, while sacrificing some information granularity, is crucial for downstream tasks in autonomous systems such as prediction and planning. Quantitative evaluation using various performance and calibration metrics showed that our proposed HMC learning rule achieved not only improved classification performances but also surpassed the baselines in terms of calibration and robustness against out-of-distribution data, all while demanding relatively few computational resources.

Future research should explore whether imposing more structured prior knowledge derived from human scene understanding enhances model performance, or allowing the model to autonomously learn flexible structures to abstract representations. This exploration could provide valuable insights into domain generalization, as well as semi- or unsupervised 3D semantic segmentation.
\FloatBarrier


\addtolength{\textheight}{-12cm}   


\bibliographystyle{IEEEtran}
\bibliography{mybibfile}

\end{document}